%% file: main.tex
\documentclass[10pt,twocolumn,letterpaper]{article}

\usepackage[algorithms]{wacv}      % To produce the REVIEW version for the algorithms track

\definecolor{wacvblue}{rgb}{0.21,0.49,0.74}
\usepackage[pagebackref,breaklinks,colorlinks,allcolors=wacvblue]{hyperref}

\usepackage{colortbl}
\definecolor{lightred}{RGB}{255, 220, 220}
\definecolor{lightyellow}{rgb}{1.0, 1.0, 0.8}
\definecolor{lightgreen}{RGB}{215,233,218}

\usepackage{balance}
\usepackage{pifont}
\newcommand{\cmark}{\textcolor{green!60!black}{\ding{51}}}
\newcommand{\xmark}{\textcolor{red}{\ding{55}}}

\def\wacvPaperID{1025} % *** Enter the WACV Paper ID here
\def\confName{WACV}
\def\confYear{2027}

\title{HSA: Hierarchical Slot Attention for Multi-granularity Scene-Decomposition}

\author{
\textbf{Neelu Madan}$^{1,2}$, \quad 
\textbf{Rongzhen Zhao}$^{3}$, \quad 
\textbf{Andreas Mogelmose}$^{1,2}$, \quad
\textbf{Juho Kannala}$^{3}$ \\
\textbf{Joni Pajarinen}$^{3}$ \quad
\textbf{Graham W. Taylor}$^{4,5}$, \quad 
\textbf{Thomas B. Moeslund}$^{1,2}$\\
$^{1}$Aalborg University, Denmark
$^{2}$Pioneer Centre for AI, Denmark,
$^{3}$Aalto University, Finland, \\
$^{4}$University of Guelph, Canada, 
$^{5}$Vector Institute, Canada \\
}

\begin{document}
\maketitle
\vspace*{-0.4cm}

\begin{abstract}

Slot attention is a powerful framework for object-centric learning, decomposing visual scenes into latent slots through iterative competitive attention. However, existing methods share two critical limitations: they decompose scenes into a flat set of slots at a single granularity, and this decomposition is based on appearance rather than semantics. Yet humans understand scenes through semantic hierarchies: separating foreground from background, recognizing object categories, and identifying individual instances. Crucially, such semantic hierarchies cannot emerge without supervision, because category names are human constructs, not visual patterns. We propose Hierarchical Slot Attention (HSA), which learns multi-granularity semantic scene decomposition from a single model. HSA decomposes scenes at three levels: holistic (foreground/background), semantic (object categories), and panoptic (individual instances). Using only 10\% labeled data, combined with hierarchical alignment loss, HSA learns all three levels jointly. We further introduce grouping purity and containment to measure whether the hierarchy is encoded in representation space, not just output masks. Experiments on COCO and PASCAL VOC demonstrate that HSA outperforms the strongest flat baseline by up to \textbf{$+$41.5} ARI at holistic, \textbf{$+$14.6} at semantic, and \textbf{$+$10.4} at panoptic level on COCO, with even larger gains on Pascal VOC, while requiring a single model instead of three. Code will be made available upon acceptance.
\vspace*{-0.3cm}
\end{abstract}

%=============================================================================
\section{Introduction}
% what is SA?
Object-centric learning~\cite{engelcke2020genesis, Burgess-arxiv-2019, Locatello-NeurIPS-2020, Greff-ICML-2019, Greff-arxiv-2020} seeks to decompose visual scenes into meaningful entity representations~\cite{spelke1990principles}. Among recent approaches, slot attention~\cite{Locatello-NeurIPS-2020, Seitzer-ICLR-2022} has demonstrated remarkable success by treating objects as latent ``slots'' that compete to explain image features through iterative attention. Owing to its simplicity, slot attention has become a prominent approach for unsupervised object discovery~\cite{Greff-arxiv-2020} and scene decomposition \cite{Eslami-NeurIPS-2016}. Recent extensions~\cite{Seitzer-ICLR-2022, kakogeorgiou2024spot, grigore2025slotmatch, Manasyan-CVPR-2025} leverage self-supervised visual features such as DINOv2~\cite{oquab2023dinov2}, which encode rich patch-level representations where spatial regions with similar appearance cluster naturally in feature space~\cite{caron2021emerging, madan2026hyperbolic}, substantially improving slot attention's applicability to real-world scenes.

% gap?
Yet humans understand scenes not only through appearance but also through semantic meaning: a region is not just a set of pixels, it is a \textit{car}, a \textit{person}, or \textit{foreground}. This semantic understanding is inherently hierarchical and cannot emerge from appearance alone. Humans naturally organize visual information at multiple levels of abstraction: distinguishing background from foreground (holistic level), recognizing object categories (semantic level), and identifying individual instances within those categories (panoptic level)~\cite{spelke1990principles}. Symbolic cognition theory~\cite{whitehead1928symbolism, jia2023improving} and concept grounding methods~\cite{radford2021learning} root this connection between perception and naming, pointing towards a minimal grounding signal as the mechanism for understanding semantics. However, existing slot-based methods decompose scenes at only a single granularity, and this decomposition is spatial, finding regions and parts through appearance cues alone. We refer to these methods as \textit{flat} baselines in later sections. Although some works attempt hierarchical scene decomposition~\cite{kosiorek2019stacked, hinton2023represent, li2022deep}, all existing unsupervised approaches \cite{kosiorek2019stacked, hinton2023represent} still decompose scenes through spatially-guided appearance. We argue that semantic hierarchy cannot emerge without supervision, because category names are human constructs, not visual patterns. We therefore aim to extend unsupervised methods with minimal supervision for semantic scene decomposition.

% transitioning 
Prior slot attention methods exploit DINOv2's latent structure only as a reconstruction target~\cite{Seitzer-ICLR-2022, kakogeorgiou2024spot}, which grounds slots in appearance but not semantic meaning. Specifically, DINO cannot group \textit{car}$_1$ and \textit{car}$_2$ under a shared \textit{vehicle} concept without label supervision. This motivates our approach: we build on slot attention's spatial grouping, enabled by DINO features, and add a minimal concept grounding signal~\cite{whitehead1928symbolism, radford2021learning, jia2023improving} to bridge the gap between spatial and semantic hierarchy.

% how we solve? 
We propose \textbf{Hierarchical Slot Attention (HSA)}, which learns multi-granularity scene decomposition through a combination of minimal supervision and a hierarchical regularization loss. Our proposed method emphasizes encoding the semantic hierarchical structure. We employ three parallel slot attention modules on shared image features, guided by two complementary signals: (1) \textit{segmentation supervision} on only 10\% of training data and (2) \textit{hierarchical alignment} that encourages cross-level embedding consistency. This allows a \textbf{single model} to provide decompositions at holistic, semantic, and panoptic granularities without separate training per level. On COCO, HSA achieves $+$41.5, $+$14.6, and $+$10.4 ARI gains at holistic, semantic, and panoptic levels over the strongest flat baseline, with even larger gains on Pascal VOC, while requiring a single model and forward pass instead of three. In summary, our contributions are three-fold:

\begin{itemize}
    \item We demonstrate that multi-granularity scene decomposition can be learned from minimal supervision. We further introduce a hierarchical regularization loss that encourages semantic inter-level consistency.
    \item We achieve state-of-the-art results on real-world object discovery, instance segmentation, and recognition benchmarks (COCO, Pascal VOC), \textit{outperforming} fully unsupervised, and semi-supervised flat models while requiring a \textit{single shared model instead of three separate models}.
    \item We introduce grouping purity and attention containment as direct measures of semantic hierarchy, evaluating whether slot groupings and attention maps align with human-defined category structure across holistic, semantic, and panoptic levels, and show that HSA produces semantically coherent multi-level decompositions.
\end{itemize}

%=============================================================================
\section{Related Work}
\label{sec:related}

\noindent\textbf{Object-Centric Learning (OCL).}
Object-centric learning aims to decompose scenes into constituent object-level representations without direct supervision. Early approaches such as MONet~\cite{Burgess-arxiv-2019}, IODINE~\cite{Greff-ICML-2019}, and GENESIS~\cite{engelcke2020genesis} employed Variational Autoencoders \cite{kingma2014auto} to model scenes as mixtures of component distributions. 
Slot Attention~\cite{Locatello-NeurIPS-2020} introduced competitive iterative attention for object binding, later extended to real-world scenes using DINO features~\cite{caron2021emerging, oquab2023dinov2} as reconstruction targets~\cite{Seitzer-ICLR-2022}. 
Subsequent variants improve binding through autoregressive decoders~\cite{singh2022illiterate, kakogeorgiou2024spot, zhao2025dias}, denoising objectives~\cite{wu2023slotdiffusion}, learnable slot initialization~\cite{Jia-ICLR-2023}, and adaptive slot queries~\cite{gao2023enhancing, fan2024adaptive, liu2025metaslot}. 
Inductive biases from motion, depth and temporal dynamics extend slot attention to videos~\cite{kipf-ICLR-2022, Elsayed-NeurIPS-2022, Zadaianchuk-NeurIPS-2023, Manasyan-CVPR-2025, zhao2026rsfq, zhao2026ssav} and 3D scenes~\cite{yu2022unsupervised, sajjadi2022object, smith2023unsupervised, liu2024slotlifter}. 
Despite this progress, all existing methods decompose scenes at a \textit{single} granularity, which results in a flat set of slots with no hierarchical structure.

% mult-modal tasks~\cite{didolkar2025ctrl,didolkar2025ctrl}, 
% learning general-purpose object-centric representations~\cite{Didolkar2024ZeroShotOCRL, zhao2025vector}.

\noindent\textbf{Hierarchical Scene Decomposition.}
Learning visual hierarchies \cite{sabour2017dynamic, sharon2006hierarchy, hinton2023represent, xu2022groupvit, wang2025learning} has been studied from multiple perspectives. 
Classical hierarchical segmentation methods build spatial hierarchies bottom-up through agglomerative 
clustering~\cite{sharon2006hierarchy, arbelaez2010contour}, merging fine regions into coarser ones by intensity or feature similarity.
Capsule Networks~\cite{sabour2017dynamic, hinton2018matrix} capture spatial part-whole relationships through hierarchical routing, with Stacked Capsule Autoencoders~\cite{kosiorek2019stacked} demonstrating unsupervised emergence of such hierarchies. 
GLOM~\cite{hinton2023represent} proposed islands of identical embeddings to represent parse trees, while PSGNet~\cite{bear2020learning} learns hierarchical scene graphs via reconstruction through a structured bottleneck.
Transformer-based methods have explored hierarchical grouping~\cite{xu2022groupvit} and hyperbolic representations~\cite{kwon2024improving, wang2025learning, madan2026hyperbolic} for capturing part-whole structure.
%CASA~\cite{casa} learns spatial part-whole hierarchy alongside recognition without supervision, but decomposes scenes geometrically rather than semantically. 
Existing unsupervised hierarchical methods decompose scenes in spatial space, grouping regions by appearance similarity and part-whole structure. 
However, they do not recover semantic hierarchies aligned with human concepts such as categories or instances. We argue that bridging this gap likely requires semantic supervision, since category structure is not fully identifiable from appearance alone.

\noindent\textbf{Supervision and Inductive Biases in OCL.}
Reconstruction objectives alone are often under-constrained, producing appearance-coherent but semantically arbitrary decompositions. Prior work mitigates this through inductive biases at different abstraction levels, including motion and depth cues for object discovery~\cite{kipf-ICLR-2022, Elsayed-NeurIPS-2022}, spatial localization priors~\cite{kim2023shepherding}, and shape priors for 3D scene decomposition~\cite{elich2022weakly}. At a higher semantic level, language grounds slot representations through contrastive conditioning~\cite{didolkar2025ctrl} or neural-symbolic association~\cite{wang2021language}, and structured object properties provide grounding via contrastive objectives~\cite{dedhia2025neural}.
However, these approaches provide supervision at a single decomposition level. We instead apply categorical segmentation supervision across three granularity levels simultaneously, providing the minimal semantic vocabulary sufficient for grounded hierarchical decomposition.

\begin{figure*}[t]
\centering
\includegraphics[width=0.90\textwidth]{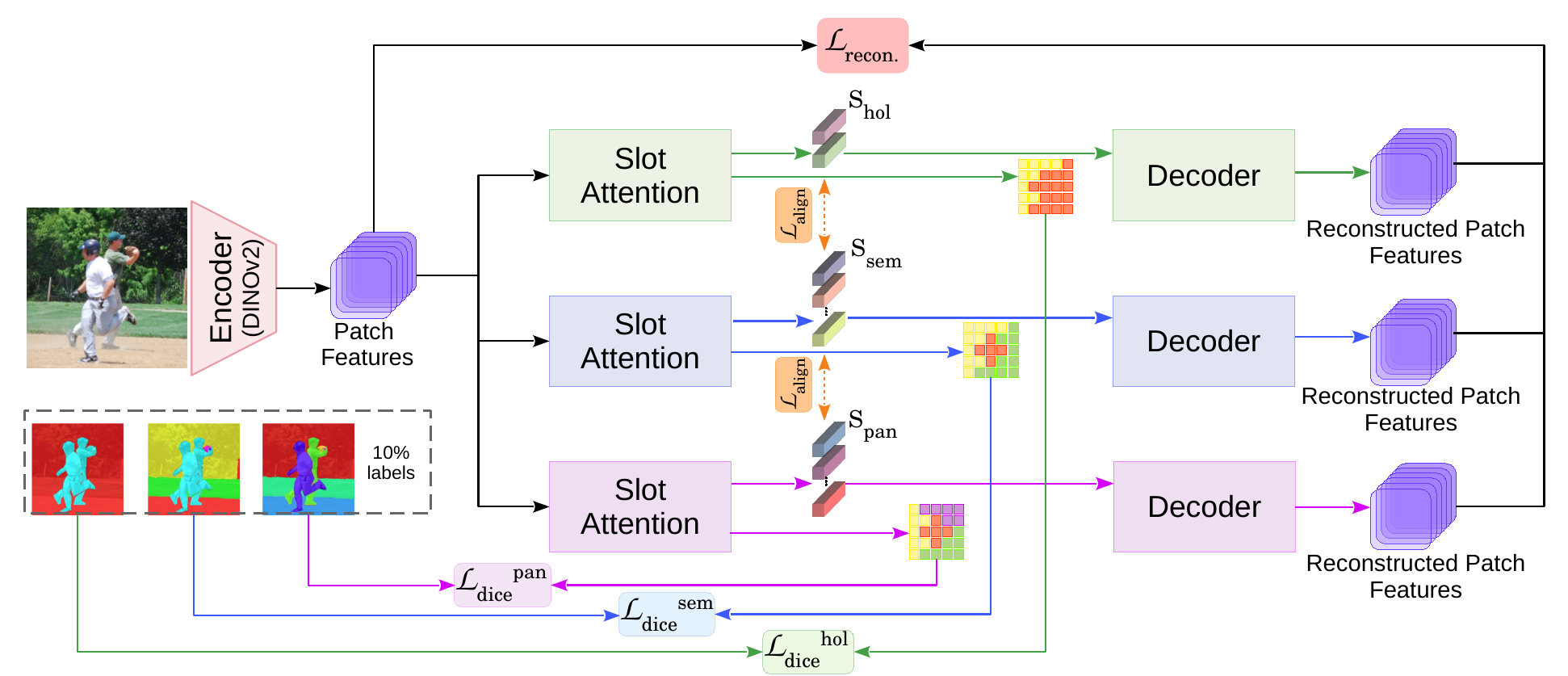}
\vspace{-0.2cm}
\caption{Hierarchical Slot Attention(HSA) shares DINOv2 patch features, which feed three level-specific slot attention modules producing $S_\text{hol}$, $S_\text{sem}$, and $S_\text{pan}$ slots. $\mathcal{L}_\textbf{align}$ enforces cross-level embedding consistency. Each level reconstructs patch features via an independent decoder; Dice losses supervise attention maps against GT masks at each granularity.}
\vspace{-0.4cm}
\label{fig:architecture}
\end{figure*}

\vspace{-0.2cm}
\section{Method}
\label{method}

\subsection{Background}
\vspace{-0.1cm}
Slot Attention~\cite{Locatello-NeurIPS-2020} decomposes visual scenes into $K$ object-centric slots through iterative competitive attention. Given image features $\mathbf{x} \in \mathbb{R}^{N \times D}$, it learns slot vectors $\mathbf{S} = \{\mathbf{s}_1, \ldots, \mathbf{s}_K\} \in \mathbb{R}^{K \times D_\text{slot}}$. At each iteration, slots act as queries competing for input features via attention weights:
\vspace{-0.2cm}
\begin{equation}
\text{attn}_{k,n} = \frac{\exp(M_{k,n})}
                    {\sum_{l=1}^{K} \exp(M_{l,n})}, \quad 
                    M_{k,n} = \frac{1}{\sqrt{D}} 
                    \mathbf{k}(\mathbf{x}_n) \cdot \mathbf{q}(\mathbf{s}_k)^T
\end{equation}
where $k \in \{1,\ldots,K\}$ indexes slots, $n \in \{1,\ldots,N\}$ indexes spatial positions, and the softmax is applied over slots. This enforces competition where each spatial position attends to exactly one slot. Slots are iteratively refined via weighted aggregation of attended features over $T$ iterations.

\subsection{Overview}
We propose \textbf{Hierarchical Slot Attention (HSA)}, which jointly learns scene decomposition at three natural semantic levels~\cite{spelke1990principles}: holistic ($K_\text{hol}$, foreground/background), semantic ($K_\text{sem}$, categories), and panoptic ($K_\text{pan}$, instances). Our core insight is that a true semantic hierarchy requires grounding via a minimal label vocabulary rather than relying solely on appearance reconstruction. \textit{To bridge these granularities, we introduce a cross-level hierarchical regularization that explicitly enforces structural coherence and embedding consistency across the emergent semantic abstraction layers.} 
As illustrated in Figure~\ref{fig:architecture}, HSA processes shared DINOv2 features through independent, level-specific slot attention modules producing $S_\text{hol}$, $S_\text{sem}$, and $S_\text{pan}$ slots. The model is optimized via a joint objective combining feature reconstruction ($\mathcal{L}_\text{recon}$), cross-level hierarchical alignment ($\mathcal{L}_\text{align}$), and categorical segmentation supervision ($\mathcal{L}_\text{sup} = \mathcal{L}_\text{dice}^\text{hol} + \mathcal{L}_\text{dice}^\text{sem} + \mathcal{L}_\text{dice}^\text{pan}$) applied to a small subset of the training data.

\subsection{Architecture}
\noindent\textbf{Shared Backbone.}
All levels share a frozen DINOv2 ViT-S/14 backbone~\cite{oquab2023dinov2} extracting patch features $\mathbf{F}_\text{backbone} \in \mathbb{R}^{H \times W \times 384}$, which are projected to a 256-dimensional space $\mathbf{F} \in \mathbb{R}^{H \times W \times D}$ ($D=256$) via a two-layer pre-norm MLP.

\noindent\textbf{Level-Specific Slot Attention.}
Instead of a single shared aggregator, HSA uses three independent modules $\{\text{SlotAttention}_i\}_{i \in \{\text{hol, sem, pan}\}}$ operating on the same backbone features:
\begin{equation}
\mathbf{S}^{(i)}, \mathbf{A}^{(i)} = \text{SA}_i\!\left(\mathbf{F},\, \mathbf{Q}^{(i)},\, K_i,\, T{=}3\right)
\end{equation}
where $\mathbf{S}^{(i)} \in \mathbb{R}^{K_i \times D}$ are slot embeddings and $\mathbf{A}^{(i)} \in \mathbb{R}^{K_i \times H \times W}$ are final-iteration attention maps normalized over slots~\cite{Seitzer-ICLR-2022, kakogeorgiou2024spot}, serving as our predicted segmentation masks. Queries are initialized from learnable distributions $\mathbf{Q}^{(i)} \sim \mathcal{N} (\boldsymbol{\mu}^{(i)}, \boldsymbol{\sigma}^{(i)})$. This decoupled design eliminates gradient conflict across granularities, allowing specialized decomposition (validated in Table~\ref{tab:aggregator_ablation}).

\noindent\textbf{Independent Decoders.}
To reconstruct features without cross-level interference, each level utilizes an independent Transformer decoder (4 layers, 4 heads, $d{=}384$) following the autoregressive design of~\cite{kakogeorgiou2024spot}:
\begin{equation}
\hat{\mathbf{F}}^{(i)} = \text{Dec}_i(\mathbf{S}^{(i)}, \mathbf{F})
\end{equation}

\subsection{Training Objective}
The total objective balances feature reconstruction with our supervision and alignment priors:
\begin{equation}
\mathcal{L} = \mathcal{L}_\text{recon} + \lambda_\text{sup}\,\mathcal{L}_\text{sup} + \lambda_\text{align}\,\mathcal{L}_\text{align}
\end{equation}

\noindent\textbf{Feature Reconstruction.}
Following~\cite{Seitzer-ICLR-2022}, each level decodes slot embeddings to reconstruct frozen self-supervised DINOv2~\cite{oquab2023dinov2} patch features $\mathbf{F}$ rather than raw pixels, ensuring a semantically rich representation space. The loss is the total squared $L_2$ error across all levels:
\begin{equation}
\mathcal{L}_\text{recon} = \sum_{i \in \{\text{hol, sem, pan}\}} \left\|\mathbf{F} - \hat{\mathbf{F}}^{(i)}\right\|^2
\end{equation}
This objective regularizes slots across all granularities to retain complete scene information, preventing representation collapse during supervised training.

\noindent\textbf{Segmentation Supervision.}
For labeled samples, slot attention maps are supervised against ground-truth masks via a Hungarian-matched Dice loss. Specifically, the Hungarian algorithm pairs each predicted slot mask $\mathbf{A}_k$ with its optimal ground-truth counterpart $\mathbf{G}_g$ by minimizing the total pairwise distance. For a successfully matched pair $(\mathbf{A}, \mathbf{G})$, the Dice loss over spatial locations $n$ is defined as:
\vspace{-0.2cm}
\begin{equation}
    \mathcal{L}_\text{dice}(\mathbf{A}, \mathbf{G}) = 1 - \frac{2\sum_{n=1}^N A_n G_n}{\sum_{n=1}^N A_n + \sum_{n=1}^N G_n}
\vspace{-0.2cm}
\end{equation}
The total supervision loss $\mathcal{L}_\text{sup}$ sums these matched pairs across all hierarchy levels ($\mathcal{L}_\text{dice}^\text{hol} + \mathcal{L}_\text{dice}^\text{sem} + \mathcal{L}_\text{dice}^\text{pan}$). To let unsupervised reconstruction stabilize first, $\mathcal{L}_\text{sup}$ is delayed for 20\text{k} steps and then annealed to $\lambda_\text{sup} = 0.5$ (see supplementary material for empirical analysis).

\noindent\textbf{Hierarchical Alignment.}
To encourage cross-level embedding consistency, we compute pairwise cosine similarity between consecutive levels. For an adjacent level pair $(u, v) \in \{(\text{hol}, \text{sem}), (\text{sem}, \text{pan})\}$, where $u$ is the coarser and $v$ the finer level, the affinity matrix $\mathbf{C} \in \mathbb{R}^{K_u \times K_v}$ is defined as:
\begin{equation}
    \mathbf{C}_{ij} = \frac{\mathbf{s}_i^{u} \cdot \mathbf{s}_j^{v}}{\|\mathbf{s}_i^{u}\|\|\mathbf{s}_j^{v}\|}
\end{equation}
where $\mathbf{s}_i^{u} \in \mathbf{S}^{(u)}$ and $\mathbf{s}_j^{v} \in \mathbf{S}^{(v)}$. A differentiable binary assignment matrix $\mathbf{B} \in \{0, 1\}^{K_u \times K_v}$ is obtained via $\mathbf{B}_{ij} = \mathbb{I}(\mathbf{C}_{ij} = \max_{k} \mathbf{C}_{kj})$, where $\mathbb{I}(\cdot)$ denotes the indicator function. This maps each fine slot to its nearest coarse slot. To permit gradient flow through this non-differentiable operation, we employ a straight-through estimator (STE) during the backward pass, routing gradients exclusively through the assigned slot pairs. Using the hinge function notation $[x]_+ = \max(0, x)$, the total hierarchical alignment loss balances slot coverage and representation alignment across consecutive level pairs:
\vspace{-0.1cm}
\begin{equation}
\small
\mathcal{L}_\text{align} = \sum_{(u,v)}\sum_{i=1}^{K_u} \left( \underbrace{[1 - \textstyle\sum_j \mathbf{B}_{ij}]_+}_\text{coverage} + \frac{1}{K_u}\underbrace{[1 - \textstyle\sum_j \mathbf{B}_{ij}\cdot\mathbf{C}_{ij}]_+}_\text{alignment} \right)
\end{equation}
The total alignment and coverage loss, $\mathcal{L}_\text{align}$, is linearly annealed from 0 to $\lambda_\text{align}=1.0$ over the course of full training.

\section{Experiments}

\subsection{Experimental Setup}
\noindent
\textbf{Datasets.} We evaluate on \textbf{MS COCO 2017}~\cite{Lin-ECCV-2014} and \textbf{PASCAL VOC 2012}~\cite{everingham2010pascal} datasets. MS COCO provides diverse real-world images with multiple co-occurring objects at varying scales, making it particularly challenging for hierarchical decomposition as scenes require simultaneous understanding at holistic (background/foreground), semantic (object categories), and panoptic (individual instances) granularities. PASCAL VOC comprises images with fewer, more salient objects, offering a complementary evaluation setting. Our semi-supervised approach uses category-level segmentation masks on 10\% of training data to supervise decomposition at all three hierarchical levels (holistic, semantic, panoptic).

\noindent
\textbf{Evaluation Metrics.}
We use three metrics to evaluate segmentation quality. \textbf{Adjusted Rand Index (ARI)} measures clustering similarity between predicted and ground-truth pixel groupings. \textbf{Mean Best Overlap (mBO)}~\cite{pont2016multiscale} performs one-to-one matching between ground-truth and predicted masks via maximum IoU, then averages the matched scores, providing a strict measure that penalizes spurious predictions. \textbf{Mean Intersection-over-Union (mIoU)} computes average overlap without one-to-one constraints, allowing multiple predictions per ground-truth object for a more lenient coverage measure. All metrics are reported at holistic, semantic, and panoptic granularities.

\noindent
\textbf{Baselines.}
We compare against five representative unsupervised slot attention methods: DINOSAUR~\cite{Seitzer-ICLR-2022}, SPOT~\cite{kakogeorgiou2024spot}, SLATE~\cite{singh2022illiterate}, SlotDiffusion~\cite{wu2023slotdiffusion}, and VQDINO~\cite{zhao2025vector}. These span the main design axes of real-world OCL: reconstruction target (DINO features vs.\ discrete tokens vs.\ raw pixels), decoder architecture (transformer vs.\ diffusion), and slot initialization strategy. All baselines are fully unsupervised. For fair comparison, each baseline is trained separately at $K{=}2$, $K{=}6$, and $K{=}11$ on COCO (and $K{=}2$, $K{=}4$, $K{=}6$ on VOC), matching HSA's three granularity levels. \textbf{Each baseline therefore requires three forward passes to cover all levels, while HSA provides all three granularities from a single forward pass with a shared backbone}.

\noindent
\textbf{Implementation Details.}
All models use a frozen DINOv2 ViT-S/14 backbone~\cite{oquab2023dinov2} extracting $14{\times}14$ patch features from $224{\times}224$ images, projected via a two-layer MLP to 256-dimensional slot space. Slot Attention runs for 3 iterations with learnable per-slot Gaussian initialisation. The number of slots at each level is chosen to match the average scene complexity of each dataset, following flat baselines like DINOSAUR~\cite{Seitzer-ICLR-2022} and SPOT~\cite{kakogeorgiou2024spot}. Specifically, we set $K_\text{hol}{=}2$ (holistic), $K_\text{sem}{=}6$ (semantic), and $K_\text{pan}{=}11$ (panoptic) on COCO to mirror its average region, category, and instance counts. On Pascal VOC, the lower object density yields $K_\text{hol}{=}2$, $K_\text{sem}{=}4$, and $K_\text{pan}{=}6$. Each hierarchy level uses an independent Transformer decoder (4 layers, 4 heads, $d{=}384$).

Models are trained with Adam ($\text{lr}{=}2{\times}10^{-4}$, 5K warmup, cosine decay), batch size 32, gradient clipping 1.0, for 100K steps on COCO and 50K on VOC. The supervised Dice loss uses 10\% labeled samples, delayed for 20K steps then annealed to weight 0.5. The alignment loss is annealed from 0 to 1 over full training. Augmentation: random crop ($[0.75, 1.0]$), horizontal flip, ImageNet normalisation.

% Table 1: hierarchical results on object discovery task. 
\begin{table*}[t]
\centering
\caption{Multi-granularity decomposition on COCO val2017 and PASCAL VOC 2012. We compare flat baselines in two settings: fully unsupervised (red) and with 10\% supervision (yellow), both requiring 3 separate models. HSA (green) learns all three levels jointly from a single shared model. Best results in \textbf{bold}; second best in \textbf{underline}.}
\vspace{-0.2cm}
\label{tab:main_results}
\small
\setlength{\tabcolsep}{1.5pt}
\resizebox{\textwidth}{!}{
\begin{tabular}{|l|ccc|ccc|ccc|ccc|ccc|ccc|}
\hline
& \multicolumn{9}{c|}{\textbf{COCO}} & \multicolumn{9}{c|}{\textbf{Pascal VOC}} \\
\cline{2-19}
& \multicolumn{3}{c|}{\textbf{Hol.} ($K_\text{hol}{=}2$)}
& \multicolumn{3}{c|}{\textbf{Sem.} ($K_\text{sem}{=}6$)}
& \multicolumn{3}{c|}{\textbf{Pan.} ($K_\text{pan}{=}11$)}
& \multicolumn{3}{c|}{\textbf{Hol.} ($K_\text{hol}{=}2$)}
& \multicolumn{3}{c|}{\textbf{Sem.} ($K_\text{sem}{=}4$)}
& \multicolumn{3}{c|}{\textbf{Pan.} ($K_\text{pan}{=}6$)} \\
Method & ARI($\uparrow$) & mBO($\uparrow$) & mIoU($\uparrow$) 
       & ARI($\uparrow$) & mBO($\uparrow$) & mIoU($\uparrow$) 
       & ARI($\uparrow$) & mBO($\uparrow$) & mIoU($\uparrow$)
       & ARI($\uparrow$) & mBO($\uparrow$) & mIoU($\uparrow$) 
       & ARI($\uparrow$) & mBO($\uparrow$) & mIoU($\uparrow$) 
       & ARI($\uparrow$) & mBO($\uparrow$) & mIoU($\uparrow$) \\
\hline
\hline
\rowcolor{lightred} \multicolumn{19}{|l|}{\textit{Flat baselines (\textbf{3 separate models, 3 forward passes}, unsupervised)}} \\
\hline
DINOSAUR \cite{Seitzer-ICLR-2022} & 19.5 & 45.3 & 44.6 & 35.4 & 33.4 & 31.4 & 33.3 & 30.2 & 28.5
  & 30.8 & 53.6 & 53.1 & 20.7 & 48.3 & 48.3 & 18.7 & 41.9 & 41.7 \\
SLATE \cite{singh2022illiterate} & 16.0 & 43.5 & 42.8 & 37.0 & 33.8 & 31.8 & 33.2 & 29.9 & 28.2
  & 32.4 & 54.7 & 54.5 & 21.1 & 49.4 & 49.3 & 20.1 & 43.1 & 42.7 \\
SlotDiffusion \cite{wu2023slotdiffusion} & 23.8 & 48.1 & 47.3 & 37.8 & 34.1 & 32.0 & 33.4 & 30.1 & 28.6
  & 13.7 & 44.6 & 43.0 & 20.6 & 48.7 & 48.7 & 19.4 & 43.3 & 43.0 \\
SPOT \cite{kakogeorgiou2024spot} & 27.5 & 49.9 & 49.1 & 40.6 & 36.7 & 34.6 & 36.7 & \underline{32.9} & \underline{31.2}
  & \underline{58.5} & \underline{67.9} & \underline{67.8} & \underline{27.9} & \underline{54.5} & \underline{54.3} & \underline{21.6} & \underline{45.9} & \underline{45.6} \\
VQDINO \cite{zhao2025vector} & 29.9 & \underline{51.4} & \underline{50.9} & 16.6 & 41.5 & 41.4 & 36.1 & 31.9 & 30.3
  & 30.8 & 53.2 & 52.5 & 23.0 & 50.5 & 50.4 & 20.6 & 43.0 & 42.8 \\
\hline
\rowcolor{lightyellow} \multicolumn{19}{|l|}{\textit{Flat baselines (\textbf{3 separate models, 3 forward passes}, 10\% supervision)}} \\
\hline
DINOSAUR \cite{Seitzer-ICLR-2022} & 25.5 & 48.8 & 48.1 & 36.2 & 34.0 & 31.9 & 33.0 & 30.6 & 29.0
  & 12.8 & 26.1 & 24.7 & 14.8 & 31.3 & 30.7 & 16.8 & 31.1 & 30.7 \\
SLATE \cite{singh2022illiterate} & 9.7 & 39.7 & 39.3 & 40.0 & 35.8 & 33.8 & 32.3 & 29.8 & 28.2
  & 15.4 & 26.6 & 24.8 & 11.1 & 30.5 & 29.6 & 13.8 & 34.2 & 33.3 \\
SlotDiffusion \cite{wu2023slotdiffusion} & 11.2 & 40.7 & 40.1 & 39.2 & 34.7 & 32.5 & 32.9 & 29.8 & 28.2
  & 11.8 & 25.2 & 23.7 & 11.3 & 28.2 & 27.1 & 11.7 & 32.0 & 30.9 \\
SPOT \cite{kakogeorgiou2024spot} & \underline{52.2} & 46.8 & 37.0 & \underline{43.1} & 38.1 & 36.0 & 36.3 & 32.5 & 30.8
  & 12.4 & 38.4 & 36.0 & 15.3 & 39.5 & 37.3 & 18.1 & 41.0 & 39.0 \\
VQDINO \cite{zhao2025vector} & 23.6 & 47.9 & 47.1 & 18.8 & \underline{43.8} & \textbf{43.8} & \underline{37.1} & 31.9 & 30.1
  & 11.3 & 24.7 & 23.4 & 10.4 & 31.5 & 30.2 & 13.2 & 36.9 & 35.8 \\
\hline
\rowcolor{lightgreen} \multicolumn{19}{|l|}{\textit{Hierarchical model (\textbf{1 shared model, 1 forward pass}, 10\% supervision)}} \\
\hline
\textbf{HSA (ours)}
  & \textbf{69.0} & \textbf{70.8} & \textbf{70.6}
  & \textbf{55.2} & \textbf{44.6} & \underline{42.6}
  & \textbf{47.1} & \textbf{39.0} & \textbf{37.3}
  & \textbf{76.9} & \textbf{78.6} & \textbf{78.5}
  & \textbf{67.3} & \textbf{69.7} & \textbf{69.4}
  & \textbf{64.5} & \textbf{64.3} & \textbf{63.4} \\
\hline
\end{tabular}
}
\vspace{-0.3cm}
\end{table*}

\subsection{Main Results}

\begin{table}[h]
\centering
\caption{Effect of label budget on HSA performance (COCO val2017). Holistic and semantic performance improve monotonically with label budget, while panoptic shows a small non-monotonic variation at higher label ratios, attributed to the higher variance of instance-level matching at $K{=}11$. Higher is better; Best results in \textbf{bold}.}
\vspace{-0.1cm}
\label{tab:label_budget}
\small
\setlength{\tabcolsep}{2pt}
\begin{tabular}{|l|ccc|ccc|ccc|}
\hline
& \multicolumn{3}{c|}{\shortstack{\textbf{Holistic}\\($K_\text{hol}{=}2$)}} 
& \multicolumn{3}{c|}{\shortstack{\textbf{Semantic}\\($K_\text{sem}{=}6$)}} 
& \multicolumn{3}{c|}{\shortstack{\textbf{Panoptic}\\($K_\text{pan}{=}11$)}} \\
Labels & ARI & mBO & mIoU & ARI & mBO & mIoU & ARI & mBO & mIoU \\
\hline
\hline
1\%  & 61.3 & 66.6 & 66.4 & 50.4 & 42.1 & 39.9 & 46.3 & 38.1 & 36.4 \\
5\%  & 66.1 & 69.1 & 68.9 & 52.9 & 43.4 & 41.3 & \textbf{49.1} & \textbf{39.7} & \textbf{38.0} \\
10\% & \textbf{69.1} & \textbf{70.9} & \textbf{70.7} & \textbf{55.5} & \textbf{44.7} & \textbf{42.7} & 47.3 & 39.0 & 37.4 \\
\hline
\end{tabular}
\vspace{-0.2cm}
\end{table}

\noindent \textbf{Label Efficiency.}
Table~\ref{tab:label_budget} shows HSA's performance with 1\%, 5\%, and 10\% labeled training data. Performance improves monotonically with label budget across holistic and semantic levels, with semantic ARI gaining $+$5.1 points from 1\% to 10\%. While allocating a larger label budget (e.g., 20\%) would predictably yield further absolute performance gains, we cap our evaluation at 10\% to strictly adhere to the low-data benchmarks established in semi-supervised literature~\cite{zhai2019s4l, kim2023shepherding}. \textit{Notably, even with only 1\% labels, HSA already outperforms all unsupervised flat baselines at every granularity (Table~\ref{tab:main_results}),} demonstrating that a minimal symbolic grounding signal is sufficient to bootstrap semantically grounded hierarchical decomposition.

\noindent\textbf{Object Discovery Task.}
Table~\ref{tab:main_results} compares HSA against flat baselines under two settings: fully unsupervised and with 10\% supervision. Each baseline requires three separately trained models to cover all granularity levels, while HSA provides all three from a single forward pass. Against unsupervised flat baselines on COCO, HSA outperforms the strongest (SPOT) by large margins: $+$41.5 ARI at holistic, $+$14.6 at semantic, and $+$10.4 at panoptic. 
Notably, adding supervision to flat baselines yields inconsistent improvements. For example, SPOT gains substantially at the holistic level ($+$24.7 ARI) but degrades at panoptic ($-$0.4 ARI), while SLATE and SlotDiffusion deteriorate across multiple levels. 
\textbf{This confirms that supervision alone is insufficient without hierarchical joint training.} HSA's joint training with shared features enables each level to benefit from the others, producing coherent multi-level decompositions that flat models cannot achieve regardless of supervision. Notably, supervision at the holistic level ($K{=}2$) hurts flat models (e.g., DINOSAUR drops from 30.8 to 12.8 ARI on VOC), as independently applied categorical supervision conflicts with the reconstruction objective without cross-level semantic context. Gains are even larger on Pascal VOC, where HSA achieves 76.9\%, 67.3\%, and 64.5\% ARI compared to SPOT's best of 58.5\%, 27.9\%, and 21.6\%, reflecting VOC's  less cluttered scene structure with fewer, larger objects that allow semantic supervision to ground slot decompositions more cleanly.

\begin{table}[h]
\vspace{-0.1cm}
\centering
\caption{Instance-level object discovery on COCO val2017. Baselines are unsupervised flat models with $K{=}7$ slots. HSA uses a single model with separate level-specific aggregators and minimal supervision. Best results in \textbf{bold}.}
\vspace{-0.2cm}
\label{tab:instance_seg}
\small
\setlength{\tabcolsep}{2pt}
\begin{tabular}{|l|cccc|}
\hline
Method & ARI($\uparrow$) & FG-ARI($\uparrow$) & mBO($\uparrow$) & mIoU($\uparrow$) \\
\hline
\hline
DINOSAUR~\cite{Seitzer-ICLR-2022} & 21.1$_{\pm1.0}$ & 37.0$_{\pm1.2}$ & 28.7$_{\pm0.5}$ & 27.3$_{\pm0.5}$ \\
SLATE~\cite{singh2022illiterate} & 20.5$_{\pm0.6}$ & 28.8$_{\pm0.3}$ & 27.4$_{\pm0.3}$ & 26.1$_{\pm0.3}$ \\
SlotDiffusion~\cite{wu2023slotdiffusion} & 20.6$_{\pm0.6}$ & 29.0$_{\pm0.1}$ & 27.5$_{\pm0.4}$ & 26.1$_{\pm0.4}$ \\
VQDINO~\cite{zhao2025vector} & 24.4$_{\pm2.0}$ & 31.5$_{\pm1.1}$ & 30.2$_{\pm0.6}$ & 28.8$_{\pm0.8}$ \\
DIAS~\cite{zhao2025dias} & 25.6$_{\pm0.1}$ & 41.2$_{\pm0.3}$ & 31.7$_{\pm0.1}$ & 30.2$_{\pm0.1}$ \\
SPOT~\cite{kakogeorgiou2024spot} & -- & 37.0$_{\pm0.1}$  & 35.0$_{\pm0.1}$ & 33.0$_{\pm0.1}$ \\
\hline
\textbf{HSA (ours)} & \textbf{31.65}$_{\pm0.1}$ & \textbf{42.74}$_{\pm0.3}$ & \textbf{36.88}$_{\pm0.2}$ & \textbf{35.29}$_{\pm0.1}$ \\
\hline
\end{tabular}
\vspace{-0.2cm}
\end{table}

\noindent \textbf{Instance-Level Object Discovery.}
For fair comparison with the baseline convention ($K{=}7$ slots), we run HSA's panoptic-level slot attention with $K{=}7$ at inference and compare against numbers reported in the original publications for five unsupervised OCL methods: DINOSAUR, SLATE, SlotDiffusion, and VQDINO (reported in~\cite{zhao2025vector}), DIAS~\cite{zhao2025dias}, 
and SPOT~\cite{kakogeorgiou2024spot}. All baselines are flat models specialized for a single decomposition level, with no hierarchical supervision. HSA achieves 
the best FG-ARI, mBO, and mIoU among all compared methods (42.74, 36.88, and 35.29, respectively), outperforming the strongest baseline on each metric by 1.5--1.9 points, while simultaneously providing holistic and semantic decompositions from the same model. This demonstrates that hierarchical training with minimal supervision not only enables multi-granularity decomposition but also improves instance-level object discovery beyond what specialized single-level models achieve.
\vspace{-0.2cm}
% hierarchical tasks
\begin{table}[h]
\centering
\caption{Hierarchical structure evaluation on COCO val2017. Higher is better; Best results in \textbf{bold}.}
\vspace{-0.2cm}
\label{tab:hierarchy}
\small
\setlength{\tabcolsep}{2pt}
\begin{tabular}{|l|cc|cc|}
\hline
& \multicolumn{2}{c|}{\textbf{Containment (\%)}} & \multicolumn{2}{c|}{\textbf{Grouping Purity (\%)}} \\
Method & $L1\leftarrow L2$ & $L2\leftarrow L3$ & $L1\leftarrow L2$ & $L2\leftarrow L3$ \\
\hline
\hline
\rowcolor{lightred} \multicolumn{5}{|l|}{\textit{Flat baselines (3 separate models, 3 forward passes)}} \\
\hline
DINOSAUR~\cite{Seitzer-ICLR-2022}     & 90.6 & 81.9 & 85.9 & 88.5 \\
SLATE~\cite{singh2022illiterate}         & 90.8 & 82.6 & 83.3 & 87.5 \\
SlotDiffusion~\cite{wu2023slotdiffusion} & 88.6 & 83.4 & 87.3 & 87.3 \\
SPOT~\cite{kakogeorgiou2024spot}          & \textbf{92.5} & 86.3 & 87.2 & 87.7 \\
VQDINO~\cite{zhao2025vector} & 91.9 & 85.6 & 87.5 & 94.3 \\
\hline
\rowcolor{lightgreen} \multicolumn{5}{|l|}{\textit{Hierarchical model (1 shared model, 1 forward pass)}} \\
\hline
\textbf{HSA (ours)} & 82.7 & \textbf{86.4} & \textbf{92.75} & \textbf{94.70} \\
\hline
\end{tabular}
\vspace{-0.2cm}
\end{table}

\begin{figure*}[th]
\centering
\includegraphics[width=0.92\textwidth]{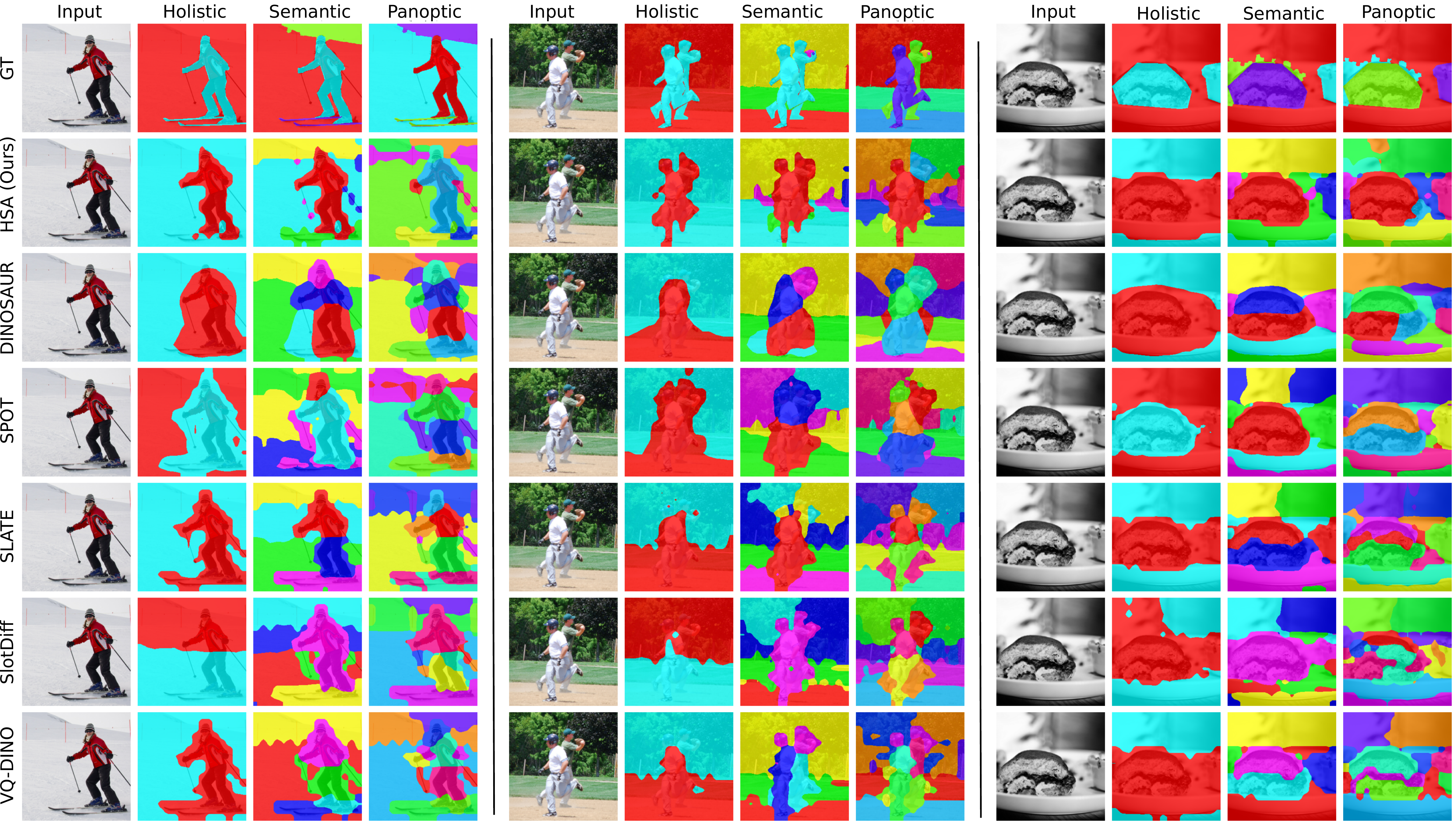}
\caption{Qualitative comparison on COCO val2017. \textbf{Row 2 (HSA, ours)}: cleanly separate objects, and are semantically grounded. \textbf{Rows 3--7}: flat baselines (DINOSAUR, SPOT, SLATE, SlotDiffusion, VQDINO) produce semantically arbitrary decompositions.  Best viewed in color.}
\vspace{-0.4cm}
\label{fig:qualitative_coco}
\end{figure*}

\paragraph{Hierarchical Structure Evaluation.}
We evaluate whether HSA learns genuine hierarchical structure across levels using two complementary metrics. \textbf{Attention containment} measures the fraction of each fine-level slot's attention mass falling within its best-matching coarse-level slot (1.0 = perfect nesting; random $\approx 1/K_{\text{parent}}$). \textbf{Grouping purity} measures whether fine-level slots assigned to the same coarse slot share the same ground-truth category (100\% = perfect semantic grouping). We evaluate flat baselines in their unsupervised setting, as independently trained models have no cross-level training signal regardless of the supervision signal. Supplementary material includes a comparison with 10\% supervised flat baselines.

Table~\ref{tab:hierarchy} shows that all methods score well on containment $L1\leftarrow L2$, confirming this transition is partly a \textit{spatial} property of slot count rather than a learned signal. Specifically, with only $K{=}2$ coarse slots, each covers a large image region and trivially contains most finer slots regardless of training. On the more discriminative $L2\leftarrow L3$ transition, HSA achieves $86.4\%$ containment, showing competitive performance with the strongest flat baseline ($86.3\%$). Crucially, HSA significantly outperforms all flat baselines in grouping purity, achieving $92.75\%$ on $L1\leftarrow L2$ (a $5.5\text{--}9.4\%$ improvement). On $L2\leftarrow L3$, HSA achieves 94.70\%, exceeding all baselines. Together, these results demonstrate that hierarchical regularization produces tighter semantic grouping and better spatial nesting than three independently trained models, using a single model. \textbf{Notably, while flat baselines encode appearance-based hierarchies, HSA encodes semantically grounded hierarchies.}

\begin{table}[h]
\centering
\caption{Object recognition on COCO val2017. Top-1 and Top-3 accuracy over Hungarian-matched slot--object pairs; \#BBox counts matched pairs (slot recall). Best results in \textbf{bold}.}
\vspace{-0.2cm}
\label{tab:recognition}
\small
\setlength{\tabcolsep}{6pt}
\begin{tabular}{|l|ccc|}
\hline
Method & Top-1 ($\uparrow$) & Top-3 ($\uparrow$) & \#BBox ($\uparrow$)\\
\hline
\hline
\rowcolor{lightred} \multicolumn{4}{|l|}{\textit{Flat baselines (3 separate models, 3 forward passes)}} \\
\hline
DINOSAUR \cite{Seitzer-ICLR-2022}        & 34.1 & 58.4 & 6842 \\
SLATE \cite{singh2022illiterate}         & 27.5 & 47.9 & 6779 \\
SPOT \cite{kakogeorgiou2024spot}         & 38.2 & 62.4 & 6992 \\
VQDINO \cite{zhao2025vector}             & 32.8 & 56.3 & 6910 \\
SlotDiffusion \cite{wu2023slotdiffusion} & 25.5 & 46.4 & 6889 \\
\hline
\rowcolor{lightgreen} \multicolumn{4}{|l|}{\textit{Hierarchical model (1 shared model, 1 forward pass)}} \\
\hline
\textbf{HSA (ours)} & \textbf{38.9} & \textbf{63.8} & \textbf{7045} \\
\hline
\end{tabular}
\vspace{-0.3cm}
\end{table}
% object-recognition task
\noindent\textbf{Object Recognition Task.}
Table~\ref{tab:recognition} evaluates slot-based object recognition on COCO val2017. For each image, predicted slots are matched to ground-truth instances via Hungarian matching; matched pairs are passed to a recognition MLP trained on slot embeddings. Top-1 and Top-3 accuracy measure whether the correct COCO category appears in the top-1 or top-3 predictions over matched pairs. \#BBox counts successfully matched slot--object pairs, reflecting slot recall rather than classification quality. We compare against unsupervised flat baselines; results with 10\% supervised flat baselines are provided in the supplementary material. HSA achieves the best Top-1 (38.9\%) and Top-3 (63.8\%) accuracy among all methods, and the \textit{highest slot recall} (\#BBox = 7045), demonstrating that semantically grounded slot representations not only improve segmentation but also produce more discriminative embeddings for downstream recognition. 

\begin{table}[h]
\centering
\caption{Model efficiency comparison. Baselines require three separate models and three forward passes for full hierarchical decomposition. HSA covers all three granularity levels in a single forward pass. Params in millions (M); inference time per image on a single GPU. Lower is better; best results in \textbf{bold}}
\vspace{-0.2cm}
\label{tab:efficiency}
\small
\setlength{\tabcolsep}{1.5pt}
\begin{tabular}{|l|cc|ccc|c|}
\hline
& \multicolumn{2}{c|}{\textbf{Params (M)}} 
& \multicolumn{4}{c|}{\textbf{Inference Time (ms)}} \\
\cline{2-7}
Method & Total & Train. & $K_\text{hol}{=}2$ & $K_\text{sem}{=}6$ & $K_\text{pan}{=}11$ & Total \\
\hline
\hline
DINOSAUR      & 99.0  & 34.2  & $11.9_{\pm1.3}$ & $18.2_{\pm1.5}$ & $25.7_{\pm0.4}$ & 55.8 \\
SPOT          & 98.4  & 33.6  & $9.9_{\pm0.1}$  & $10.0_{\pm0.2}$ & $10.1_{\pm0.1}$ & 30.0 \\
SLATE         & 92.4  & 20.7  & $17.1_{\pm0.3}$ & $17.3_{\pm0.3}$ & $17.3_{\pm0.3}$ & 51.7 \\
SlotDiff & 478.5 & 410.1 & $29.4_{\pm0.2}$ & $29.4_{\pm0.2}$ & $29.5_{\pm0.2}$ & 88.3 \\
VQDINO        & 111.3 & 20.7  & $11.6_{\pm0.3}$ & $11.9_{\pm0.9}$ & $11.8_{\pm0.8}$ & 35.3 \\
\hline
\textbf{HSA (ours)} & \textbf{52.5} & \textbf{30.9} & \multicolumn{3}{c|}{single forward pass} & \textbf{17.3}$_{\pm1.6}$ \\
\hline
\end{tabular}
\vspace{-0.2cm}
\end{table}

\noindent\textbf{Efficiency.}
Table~\ref{tab:efficiency} compares model efficiency for full three-level hierarchical decomposition. Flat baselines require three separate models (one per 
granularity level) resulting in $3\times$ parameter overhead and three sequential forward passes at inference. HSA covers all three levels in a single forward pass of $17.3_{\pm1.6}$ms, compared to 30--88ms required by flat baselines. In terms of parameters, HSA uses 52.5M total compared to $3\times$159.5M = 478.5M for SlotDiffusion. \textbf{This efficiency advantage comes without sacrificing performance, meaning HSA outperforms all flat baselines on segmentation quality at every granularity level (Table~\ref{tab:main_results})}. This demonstrates that joint hierarchical learning is both more efficient and effective than independent single-level training.

%%%%%%%%%%%%%%%%%%%%%%%%%%%%%%%%%%%%%%%%%%%%%%%%%%%%%%%%%%%%%%%%%%%%%%%%%%%%%%%%%%%%%%%%%%%%%%
\subsection{Qualitative Analysis}

Figure~\ref{fig:qualitative_coco} visualizes decompositions at all three granularity levels for five flat baselines and HSA on representative COCO images. Flat baselines (rows 3--7), trained without supervision, produce geometrically coherent but semantically arbitrary decompositions slots partition the scene into visually uniform regions without consistent correspondence to meaningful categories. In contrast, HSA (row 2) produce semantically grounded decompositions at every level: the holistic level cleanly separates foreground (skier) from background; the semantic level assigns distinct slots to object categories (person, trees, snow, sky); and the panoptic level separates individual instances. This confirms the central claim of our work, slot methods guided by minimal categorical supervision are sufficient to bridge the gap between spatial and semantic hierarchy, producing decompositions that align with human scene understanding in a way that purely unsupervised methods cannot.

\begin{figure}[h]
\vspace{-0.3cm}
\centering
\includegraphics[width=0.92\linewidth]{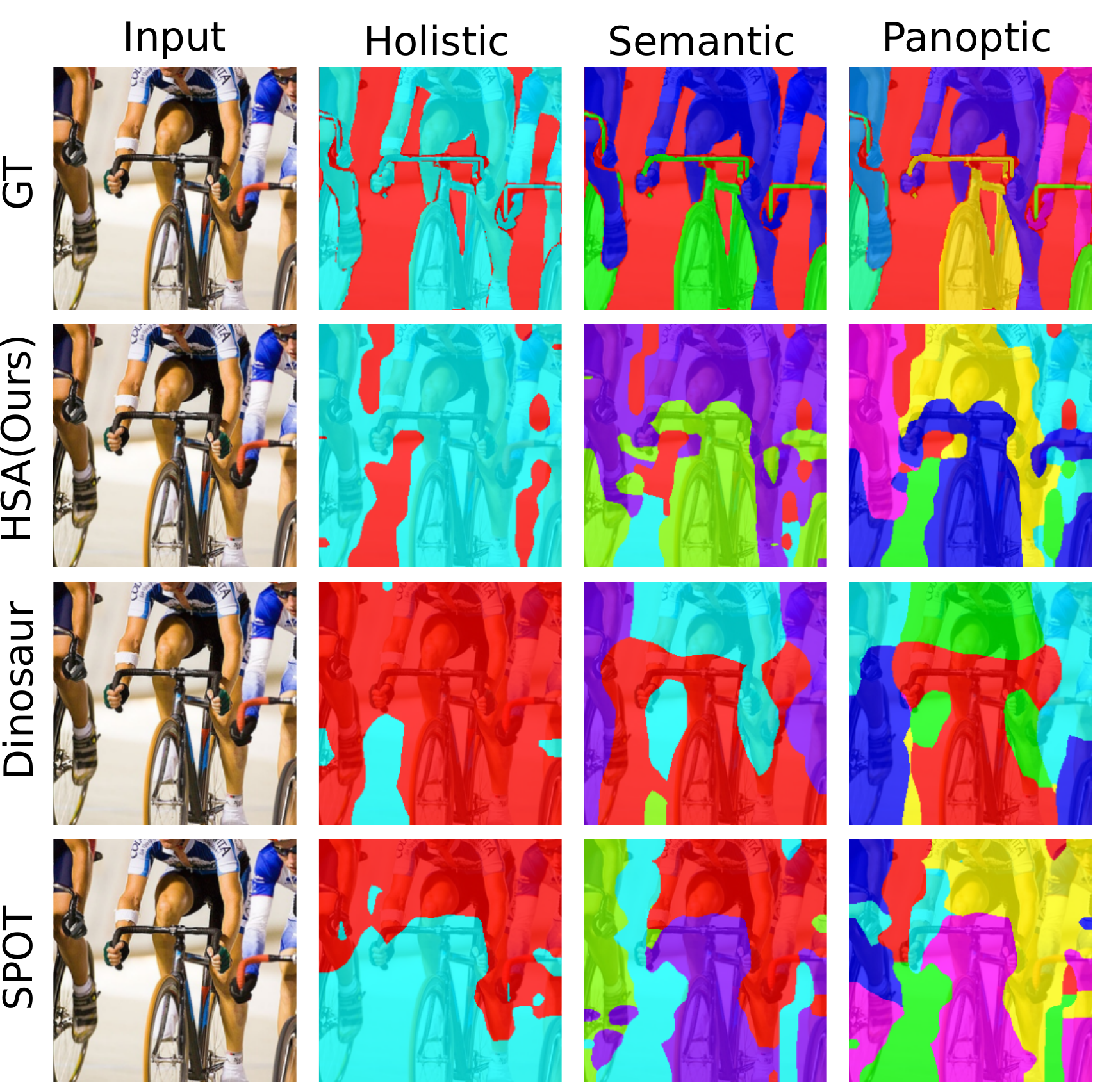}
\vspace{-0.2cm}
\caption{Qualitative results on Pascal VOC. HSA (row 1) produces cleaner part boundaries and more coherent semantic grouping than flat baselines. Best viewed in color.}
\label{fig:qualitative_voc}
\vspace{-0.3cm}
\end{figure}

Figure~\ref{fig:qualitative_voc} shows qualitative results on Pascal VOC. HSA produces cleaner part boundaries and more semantically coherent groupings at the semantic and panoptic levels compared to flat baselines, which fragment the bicycle and riders into visually similar but semantically arbitrary regions. This confirms that HSA's advantage generalizes beyond COCO to scenes with fewer, larger, more salient objects.

\noindent \textbf{Beyond Quantitative Metrics.}
Figure~\ref{fig:qualitative_coco} reveals cases where HSA's semantic and panoptic level discovers object boundaries absent from GT annotations, i.e., separating ski poles from the skier in the left examples, and the plate from surrounding food in the right example. These fine-grained, semantically valid splits are penalized by ARI, which scores any deviation from GT clustering as incorrect, regardless of whether the discovered decomposition is itself meaningful. \textit{This highlights a known limitation of standard OCL evaluation: GT annotations define a single fixed granularity per level, while real scenes often admit multiple valid decompositions that the model may correctly discover but the metric cannot reward.} We argue these points to a broader limitation in how object-centric learning is evaluated, where pixel-clustering metrics such as ARI cannot distinguish between an incorrect decomposition and a finer, equally valid one.

%%%%%%%%%%%%%%%%%%%%%%%%%%%%%%%%%%%%%%%%%%%%%%%%%%%%%%%%%%%%%%%%%%%%%%%%%%%%%%%%%%%%%%%%%%%%%%
\subsection{Ablation Studies}
\label{ablation}

\begin{table*}[t]
\centering
\caption{Ablation study on COCO val2017. We isolate the contribution of segmentation supervision (dice) and hierarchical alignment loss (align) across all three granularity levels. \cmark\ = enabled = component enabled, \xmark\ = disabled. Best per column in \textbf{bold}.}
\vspace{-0.1cm}
\label{tab:ablation1}
\setlength{\tabcolsep}{3pt}
\begin{tabular}{|l|cc|ccc|ccc|ccc|}
\hline
& & & \multicolumn{3}{c|}{\textbf{Holistic}} & \multicolumn{3}{c|}{\textbf{Semantic}} & \multicolumn{3}{c|}{\textbf{Panoptic}} \\
Config & Dice & Align & ARI ($\uparrow$) & mBO ($\uparrow$) & mIoU ($\uparrow$) & ARI ($\uparrow$) & mBO ($\uparrow$) & mIoU ($\uparrow$) & ARI ($\uparrow$) & mBO ($\uparrow$) & mIoU ($\uparrow$) \\
\hline
\hline
No dice, no align & \xmark & \xmark & 46.3 & 58.3 & 57.8 & 39.0 & 35.3 & 33.0 & 35.2 & 32.2 & 30.5 \\
No dice, align    & \xmark & \cmark & 47.1 & 58.7 & 58.1 & 39.5 & 35.5 & 33.2 & 35.6 & 32.4 & 30.7 \\
Dice, no align    & \cmark & \xmark & 66.9 & 69.6 & 69.4 & \textbf{53.5} & \textbf{44.0} & \textbf{41.9} & \textbf{49.6} & \textbf{40.2} & \textbf{38.6} \\
\textbf{Full model} & \cmark & \cmark & \textbf{67.0} & \textbf{69.6} & \textbf{69.4} & 53.5 & 43.9 & 41.8 & 49.3 & 40.0 & 38.4 \\
\hline
\end{tabular}
\end{table*}

\noindent\textbf{Component Analysis.}
Table~\ref{tab:ablation1} isolates the contribution of segmentation supervision (Dice) and hierarchical alignment across all three levels. Without any supervision or alignment (row 1), slots decompose scenes spatially but without semantic grounding. Adding alignment alone (row 2) yields marginal improvements ($+$0.4--0.8 ARI). Segmentation supervision is the dominant factor: adding Dice alone (row 3) produces large gains across all levels ($+$20.6 ARI at holistic, $+$14.5 at semantic, $+$14.4 at panoptic), demonstrating that a small symbolic grounding signal is sufficient to steer slot decompositions towards human-aligned semantic concepts. Interestingly, adding alignment on top of supervision (row 4, full model) provides negligible gains at the holistic level but slightly reduces semantic and panoptic performance. We attribute this to DINO's pretrained features already providing strong spatial alignment, and explicit alignment regularization marginally interferes with that. \textbf{This is because the supervision signal enforces hierarchies from human concepts, while DINO feature captures hierarchies from appearance or visual patterns.} This is also why we retain the alignment loss in our final model as it improves holistic decomposition and maintains cross-level embedding consistency, which is demonstrated in the grouping purity task (Table~\ref{tab:hierarchy}).

\noindent\textbf{Separate Slot Aggregator.}
Table~\ref{tab:aggregator_ablation} compares a single shared slot attention module across all three levels against level-specific aggregators. Separate aggregators improve holistic and semantic decomposition by 1.6--1.9 ARI, eliminating gradient conflict that arises when one set of weights must serve three different granularities simultaneously. The shared aggregator retains a slight advantage on panoptic ($+$2.3 ARI), likely because $K{=}11$ slots benefit from implicit feature sharing with coarser levels during joint training. We adopt separate aggregators as our main configuration, as the gains on holistic and semantic outweigh the small panoptic regression.

\begin{table}[h]
\centering
\caption{Ablation on shared vs.\ level-specific (separate) slot attention aggregators on COCO val2017. Higher is better; Best per column in \textbf{bold}.}
\label{tab:aggregator_ablation}
\vspace{-0.2cm}
\small
\setlength{\tabcolsep}{1.5pt}
\begin{tabular}{|l|ccc|ccc|ccc|}
\hline
& \multicolumn{3}{c|}{\textbf{Hol.} ($K_\text{hol}{=}2$)} 
& \multicolumn{3}{c|}{\textbf{Sem.} ($K_\text{sem}{=}6$)} 
& \multicolumn{3}{c|}{\textbf{Pan.} ($K_\text{pan}{=}11$)} \\
Config & ARI & mBO & mIoU 
     & ARI & mBO & mIoU 
     & ARI & mBO & mIoU \\
\hline
\hline
One Agg.  & 67.2 & 69.8 & 69.6 & 53.9 & 44.1 & 42.2 & \textbf{49.6} & \textbf{40.6} & \textbf{38.9} \\
Sep. Agg. & \textbf{69.1} & \textbf{70.9} & \textbf{70.7} & \textbf{55.5} & \textbf{44.7} & \textbf{42.7} & 47.3 & 39.0 & 37.4 \\
\hline
\end{tabular}
\vspace{-0.2cm}
\end{table}

\section{Conclusion}

We proposed Hierarchical Slot Attention (HSA), a shared model that learns object-centric decomposition at holistic, semantic, and panoptic levels, mirroring human scene perception~\cite{spelke1990principles}. While spatial hierarchies emerge unsupervised, we show that semantic hierarchies require minimal supervision: HSA provides this through Dice supervision on only 10\% of labeled data, while retaining the model's ability to discover objects beyond ground-truth annotations. Our hierarchical evaluation shows that baseline models already encode structure among their learned embeddings, but this structure reflects spatial and appearance correspondence rather than semantic meaning; HSA's hierarchical alignment loss instead enforces cross-level consistency among semantically grounded representations, achieving stronger alignment than baselines that encode spatial structure alone. Overall, HSA achieves significant improvements over flat baselines across object discovery, hierarchical evaluation, and recognition tasks, while requiring a shared model and a single forward pass instead of three separate models per granularity level. In future work, we aim to reduce supervision further by exploring weaker grounding signals such as language descriptions, and to extend HSA to video for temporal hierarchical decomposition.

\section{Acknowledgement}
This work was supported by the Innovation Fund Denmark, Grant No. 2081-00001B; and the Canada CIFAR AI Chairs.

{
    \small
    \bibliographystyle{ieeenat_fullname}
    \bibliography{references}
}

\newpage
\input{appendix}

\end{document}

%% file: appendix.tex
\section{Additional Experiments}

\noindent \textbf{Alignment Loss Weight Analysis.}
Table~\ref{tab:align_weight} evaluates the impact of the final alignment loss weight $\lambda_\text{align}$ on COCO val2017. We observe that setting $\lambda_\text{align} = 1.0$ provides the optimal balance for coarser granularities, yielding peak performance in both Holistic (+1.9 ARI over 0.1) and Semantic (+1.6 ARI over 0.1) levels. Interestingly, this peak at $\lambda_\text{align} = 1.0$ coincides with a localized drop in Panoptic performance, which recovers and peaks at $\lambda_\text{align} = 2.0$ (50.3 ARI). This suggests a slight optimization trade-off between strict fine-grained panoptic alignment and broader semantic clustering, though the framework remains functional across an order of magnitude of weight scales. Additionally, because our framework prioritizes balanced accuracy across all three hierarchical tiers, $\lambda_\text{align} = 1.0$ serves as the most effective parameter configuration.

\begin{table}[h]
\centering
\caption{Ablation on alignment loss weight $\lambda_\text{align}$ on COCO val2017. All metrics are higher-is-better. Best results in \textbf{bold}.}
\label{tab:align_weight}
\small
\setlength{\tabcolsep}{2.5pt}
\begin{tabular}{|l|ccc|ccc|ccc|}
\hline
& \multicolumn{3}{c|}{\textbf{Holistic}} 
& \multicolumn{3}{c|}{\textbf{Semantic}} 
& \multicolumn{3}{c|}{\textbf{Panoptic}} \\
$\lambda_\text{align}$ & ARI & mBO & mIoU 
                       & ARI & mBO & mIoU 
                       & ARI & mBO & mIoU \\
\hline
\hline
0.1 & 67.1 & 69.7 & 69.5 & 53.6 & 43.8 & 41.7 & 49.9 & 40.2 & 38.5 \\
0.5 & 67.3 & 69.7 & 69.5 & 53.5 & 43.7 & 41.6 & 49.9 & 40.2 & 38.5 \\
1.0 & \textbf{69.0} & \textbf{70.8} & \textbf{70.6} & \textbf{55.2} & \textbf{44.6} & \textbf{42.6} & 47.1 & 39.0 & 37.3 \\
2.0 & 67.1 & 69.6 & 69.4 & 53.6 & 43.8 & 41.8 & \textbf{50.3} & \textbf{40.3} & \textbf{38.6} \\
\hline
\end{tabular}
\end{table}

\begin{table}[h]
\centering
\caption{Ablation on components of $\mathcal{L}_\text{align}$ on COCO val2017. All metrics are higher-is-better. Best results in \textbf{bold}.}
\label{tab:align_components}
\small
\setlength{\tabcolsep}{1pt}
\begin{tabular}{|l|ccc|ccc|ccc|}
\hline
& \multicolumn{3}{c|}{\textbf{Holistic}} 
& \multicolumn{3}{c|}{\textbf{Semantic}} 
& \multicolumn{3}{c|}{\textbf{Panoptic}} \\
Config & ARI & mBO & mIoU 
       & ARI & mBO & mIoU 
       & ARI & mBO & mIoU \\
\hline
\hline
No align          & 67.3 & 69.7 & 69.5 & 53.5 & 43.8 & 41.7 & \textbf{49.2} & 40.3 & 38.6 \\
No coverage   & 67.0 & 69.5 & 69.3 & 53.6 & 43.9 & 41.8 & 49.0 & 40.3 & \textbf{38.6} \\
Full $\mathcal{L}_\text{align}$ & \textbf{69.0} & \textbf{70.8} & \textbf{70.6} & \textbf{55.2} & \textbf{44.6} & \textbf{42.6} & 47.1 & 39.0 & 37.3 \\
\hline
\end{tabular}
\end{table}

\noindent \textbf{Alignment Loss Component Analysis.}
Table~\ref{tab:align_components} evaluates the individual components of $\mathcal{L}_\text{align}$. Incorporating the full loss formulation (combining both alignment and coverage terms) proves highly effective for coarser levels, yielding a substantial boost of up to +2.0 ARI in Holistic and +1.7 ARI in Semantic segmentation over ablated variants. For the panoptic level ($K_\text{pan}{=}11$), removing the alignment term performs better by 2.1 ARI, suggesting that strict cross-level constraints over-compete with the larger number of panoptic slots. Specifically, as slot count increases, inter-slot competition intensifies, making fine-grained instance boundaries harder to maintain under hierarchical alignment. Tuning solely based on panoptic performance would therefore disrupt the broader hierarchical structure. We adopt the full $\mathcal{L}_\text{align}$ as it provides the best balance across holistic and semantic levels, while panoptic slots benefit from partial freedom to discover individual instances.

\begin{table}[h]
\centering
\caption{Object recognition on COCO val2017 with 10\% supervision applied to flat baselines. Adding supervision yields comparable or lower performance, which confirms adding alignment with supervision improves the performance. All metrics are higher-is-better. Best results in \textbf{bold}.}
\label{tab:recognition_sup}
\small

\setlength{\tabcolsep}{6pt}
\begin{tabular}{|l|ccc|}
\hline
Method & Top-1 ($\uparrow$) & Top-3 ($\uparrow$) & \#BBox ($\uparrow$)\\
\hline
\hline
\rowcolor{lightyellow} \multicolumn{4}{|l|}{\textit{Flat baselines (3 Seperate Models, 3 forward passes)}} \\
\hline
DINOSAUR \cite{Seitzer-ICLR-2022}        & 33.7 & 58.1 & 6804 \\
SLATE \cite{singh2022illiterate}         & 28.8 & 50.1 & 6859 \\
SPOT \cite{kakogeorgiou2024spot}         & 37.5 & 62.0 & 6949 \\
VQDINO \cite{zhao2025vector}             & 33.1 & 56.5 & 6821 \\
SlotDiffusion \cite{wu2023slotdiffusion} & 25.3 & 46.7 & 6916 \\
\hline
\rowcolor{lightgreen} \multicolumn{4}{|l|}{\textit{HSA (1 shared model, 1 forward pass)}} \\
\hline
\textbf{HSA (ours)} & \textbf{38.9} & \textbf{63.8} & \textbf{7045} \\
\hline
\end{tabular}
\end{table}

\noindent \textbf{Object Recognition: Supervised Flat Baseline Comparison}
Table~\ref{tab:recognition_sup} extends the object recognition evaluation by equipping flat baselines with a 10\% supervision budget. We noticed that introducing supervision results in equivalent or worse performance on the flat baselines (Table 1 of the main paper). We observe a similar trend on the object recognition task. Introducing labels yields comparable or even degraded performance across all baselines; for instance, SPOT's Top-1 accuracy drops from 38.2\% to 37.5\% despite receiving the same supervision budget as HSA. This counter-intuitive result confirms that supervision alone, when applied to flat slot representations, fails to enhance slot discriminability and may even introduce optimization bottlenecks. In contrast, HSA outperforms all supervised flat baselines across every metric, demonstrating that joint hierarchical training is the critical mechanism driving high-quality, discriminative slot representations.

\noindent \textbf{Hierarchical Evaluation with Supervised Flat Baselines.}
Table~\ref{tab:hierarchy_sup} extends the hierarchical structure evaluation from the main paper (Table~4) by including flat baselines trained with 10\% labeled data. Performance remains comparable to their unsupervised counterparts, confirming that independently applied supervision does not inherently improve cross-level hierarchical relations. 

Notably, while flat baselines like SPOT achieve high containment scores at the coarsest transition ($L1 \leftarrow L2$, 99.3\%). This is an artefact of the holistic Dice loss forcing $K{=}2$ slots to tightly cover the entire foreground, which trivially contains most semantic slots regardless of their placement. This 
inflated containment does not reflect genuine hierarchical structure, as confirmed by SPOT's low grouping purity at $L1\leftarrow L2$ (78.1\%). In contrast, HSA enforces a structurally disciplined decomposition, leading to significantly higher grouping purity across all levels (+4.7\% in $L1 \leftarrow L2$ and +0.3\% in $L2 \leftarrow L3$ over the best baselines). This demonstrates that a explicit cross-level training signal—rather than supervision alone—is necessary for semantically grounded hierarchical decomposition.

\begin{table}[h]
\centering
\caption{Hierarchical structure evaluation with 10\% supervised flat baselines on COCO val2017. Results are comparable to unsupervised variants (Table~4, main paper), confirming that supervision without joint training does not improve cross-level structure. All metrics are higher-is-better. Best results in \textbf{bold}.}
\label{tab:hierarchy_sup}
\small
\setlength{\tabcolsep}{2pt}
\begin{tabular}{|l|cc|cc|}
\hline
& \multicolumn{2}{c|}{\textbf{Containment (\%)}} 
& \multicolumn{2}{c|}{\textbf{Grouping Purity (\%)}} \\
Method & $L1\leftarrow L2$ & $L2\leftarrow L3$ 
       & $L1\leftarrow L2$ & $L2\leftarrow L3$ \\
\hline
\hline
\rowcolor{lightyellow} \multicolumn{5}{|l|}{\textit{Flat baselines (3 Seperate Models, 3 forward passes)}} \\
\hline
DINOSAUR      & 89.9 & 81.3 & \underline{88.1} & 88.1 \\
SLATE         & 88.0 & 80.1 & 83.2 & 88.2 \\
SlotDiffusion & 86.7 & 78.8 & 82.6 & 87.9 \\
SPOT          & \textbf{99.3} & \underline{85.2} & 78.1 & 87.9 \\
VQDINO        & \underline{91.4} & 84.0 & 87.0 & \underline{94.4} \\
\hline
\rowcolor{lightgreen} \multicolumn{5}{|l|}{\textit{HSA (1 shared model, 1 forward pass)}} \\
\hline
\textbf{HSA (ours)} & 82.7 & \textbf{86.4} 
                    & \textbf{92.8} & \textbf{94.7} \\
\hline
\end{tabular}
\end{table}

\noindent\textbf{Slot Count Ablation.}
Table~\ref{tab:slot_count} ablates the panoptic slot count $K_\text{pan}$ on COCO. Reducing $K_\text{pan}$ from 11 to 9 improves panoptic ARI by $+$6.2 points, as fewer slots reduce competition for partial object coverage. However, this comes at a small cost to holistic ($-$2.3 ARI) and semantic ($-$1.1 ARI) levels, reflecting a trade-off between panoptic specialization and cross-level feature sharing. We adopt $K_\text{pan}=11$ as our default framework setting, while noting that $K_\text{pan}=9$ may be preferable when panoptic performance is the primary objective. This insight also clarifies the non-monotonic panoptic behavior observed in our label efficiency ablation (Table 3, main paper): stronger supervision amplifies the top-down hierarchical alignment signal, i.e., benefiting holistic and semantic levels, but simultaneously intensifies slot competition when constrained to $K_\text{pan}=11$. This is just an extra ablation; \textbf{we still keep $K_\text{pan}=11$ to match the average number of panoptic objects to comply with the flat baselines~\cite{Seitzer-ICLR-2022}.}

\begin{table}[h]
\centering
\caption{Ablation on panoptic slot count $K_\text{pan}$ on COCO val2017. Reducing $K_\text{pan}$ improves panoptic ARI at a small cost to holistic and semantic levels. All metrics are higher-is-better. Best per column in \textbf{bold}.}
\label{tab:slot_count}
\small
\setlength{\tabcolsep}{2.5pt}
\begin{tabular}{|l|ccc|ccc|ccc|}
\hline
& \multicolumn{3}{c|}{\textbf{Holistic}}
& \multicolumn{3}{c|}{\textbf{Semantic}}
& \multicolumn{3}{c|}{\textbf{Panoptic}} \\
$K_\text{pan}$ & ARI & mBO & mIoU
              & ARI & mBO & mIoU
              & ARI & mBO & mIoU \\
\hline
\hline
9  & 66.7 & 69.5 & 69.3 & 54.1 & 44.0 & 41.9 & \textbf{53.5} & \textbf{40.5} & \textbf{38.5} \\
10 & 67.1 & 69.6 & 69.4 & 53.5 & 43.7 & 41.6 & 51.3 & 40.3 & 38.4 \\
11 & \textbf{69.0} & \textbf{70.8} & \textbf{70.6} & \textbf{55.2} & \textbf{44.6} & \textbf{42.6} & 47.1 & 39.0 & 37.3 \\
\hline
\end{tabular}
\end{table}

\noindent\textbf{Supervision Delay Schedule Analysis.}
Table~\ref{tab:delay_schedule} compares training with and without a 20K step delay on $\mathcal{L}_\text{sup}$. Interestingly, removing the delay altogether ("No delay") significantly improves fine-grained performance, boosting panoptic accuracy by $+$4.7 ARI. The semantic accuracy is almost equivalent in both cases. This indicates that immediate supervision helps the finer layers swiftly lock onto tight instance and category boundaries before representations consolidate. However, this early fine-grained commitment comes at a direct cost to global structural formation, causing a drop in holistic level ($-$0.9 ARI). \textbf{Because our primary framework objective is balanced cross-level hierarchical decomposition rather than optimization of a single flat layer, we retain the 20K delay as our default config}; it ensures the reconstruction engine undergoes an initial unsupervised phase change to establish stable, abstract foreground/background groupings before semantic gradients dominate. Furthermore, this aligns with our agenda of using minimal label supervision by delaying it, thereby allowing the model to focus towards the decomposition task.

\begin{table}[h]
\centering
\caption{Ablation on supervision delay schedule on COCO val2017. All metrics are higher-is-better. Best results in \textbf{bold}.}
\label{tab:delay_schedule}
\small
\setlength{\tabcolsep}{1.5pt}
\begin{tabular}{|l|ccc|ccc|ccc|}
\hline
& \multicolumn{3}{c|}{\textbf{Holistic}} 
& \multicolumn{3}{c|}{\textbf{Semantic}} 
& \multicolumn{3}{c|}{\textbf{Panoptic}} \\
Schedule & ARI & mBO & mIoU 
         & ARI & mBO & mIoU 
         & ARI & mBO & mIoU \\
\hline
\hline
No delay   & 68.1 & 70.2 & 70.0  & \textbf{55.4} & \textbf{44.9} & \textbf{42.9}  & \textbf{51.8} & \textbf{41.2} & \textbf{39.5} \\
20K (ours) & \textbf{69.0} & \textbf{70.8} & \textbf{70.6}  & 55.2 & 44.6 & 42.6  & 47.1 & 39.0 & 37.3 \\
\hline
\end{tabular}
\end{table}

\begin{table}[h]
\centering
\caption{Instance-level discovery ($K=7$ inference) using aggregator weights from each HSA hierarchy level on COCO val2017. FG-ARI, mBO, and mIoU improve monotonically from holistic to panoptic, confirming level specialization. All metrics are higher-is-better. Best results in \textbf{bold}.}
\label{tab:instance_levels}
\small
\setlength{\tabcolsep}{4pt}
\begin{tabular}{|l|cccc|}
\hline
Level & ARI & FG-ARI ($\uparrow$) & mBO ($\uparrow$) & mIoU ($\uparrow$) \\
\hline
\hline
Holistic ($K_\text{hol}=2$)  & \textbf{59.8} & 15.9 & 34.3 & 30.8 \\
Semantic ($K_\text{sem}=6$)  & 32.9          & 42.1 & 36.6 & 34.7 \\
Panoptic ($K_\text{pan}=11$) & 31.7          & \textbf{42.7} & \textbf{36.9} & \textbf{35.3} \\
\hline
\end{tabular}
\end{table}

\begin{figure*}[!t]
\centering
\includegraphics[width=0.96\textwidth]{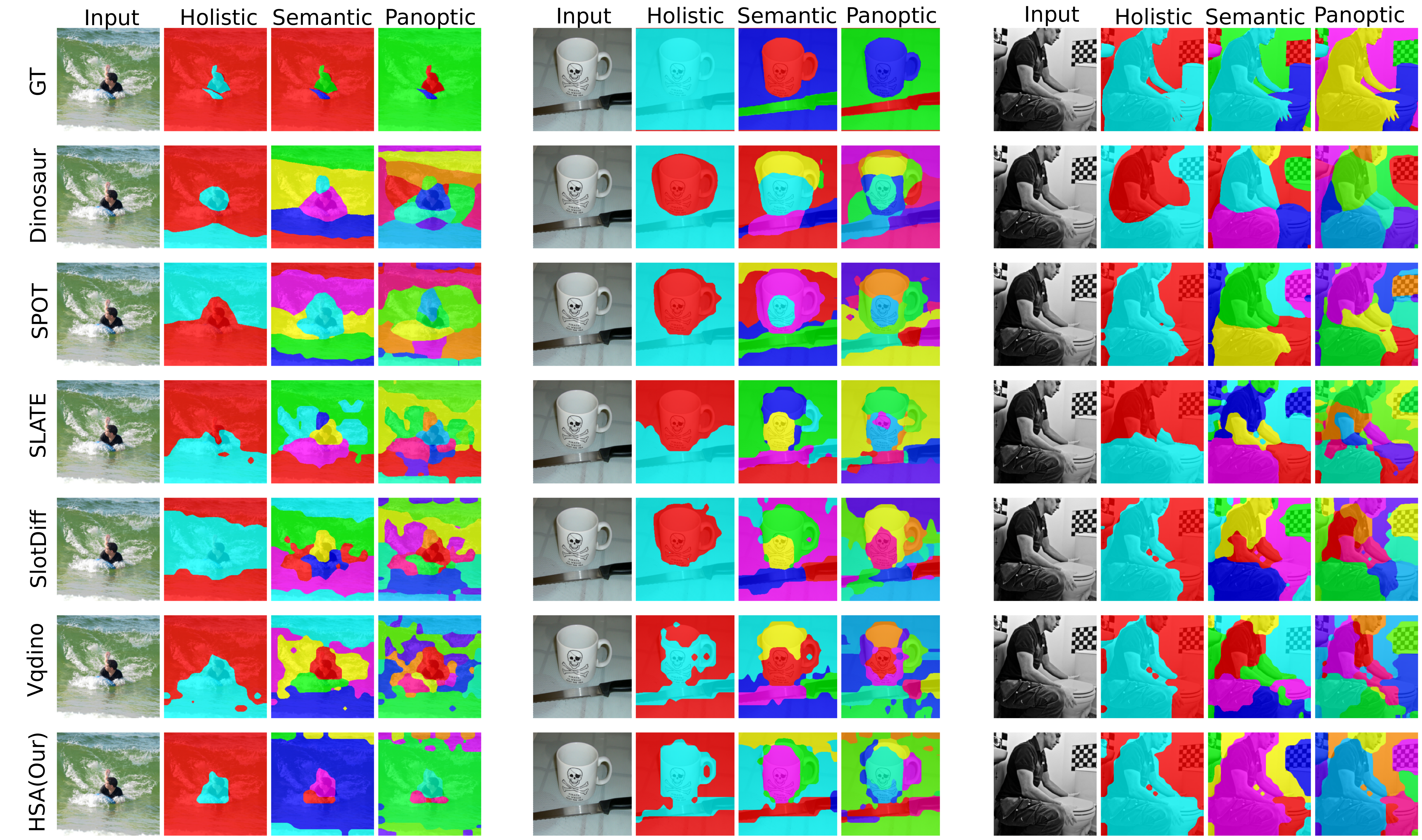}
\caption{Qualitative comparison on COCO val2017. \textbf{Rows 2--6}: flat baselines (DINOSAUR, SPOT, SLATE, SlotDiffusion, VQDINO) produce semantically arbitrary decompositions. \textbf{Row 7 (HSA, ours)}: cleanly separate objects, and are semantically grounded. Best viewed in color.}
\vspace{-0.3cm}
\label{fig:qualitative_coco_supp}
\end{figure*}

\begin{figure*}[h]
\centering
\includegraphics[width=0.99\textwidth]{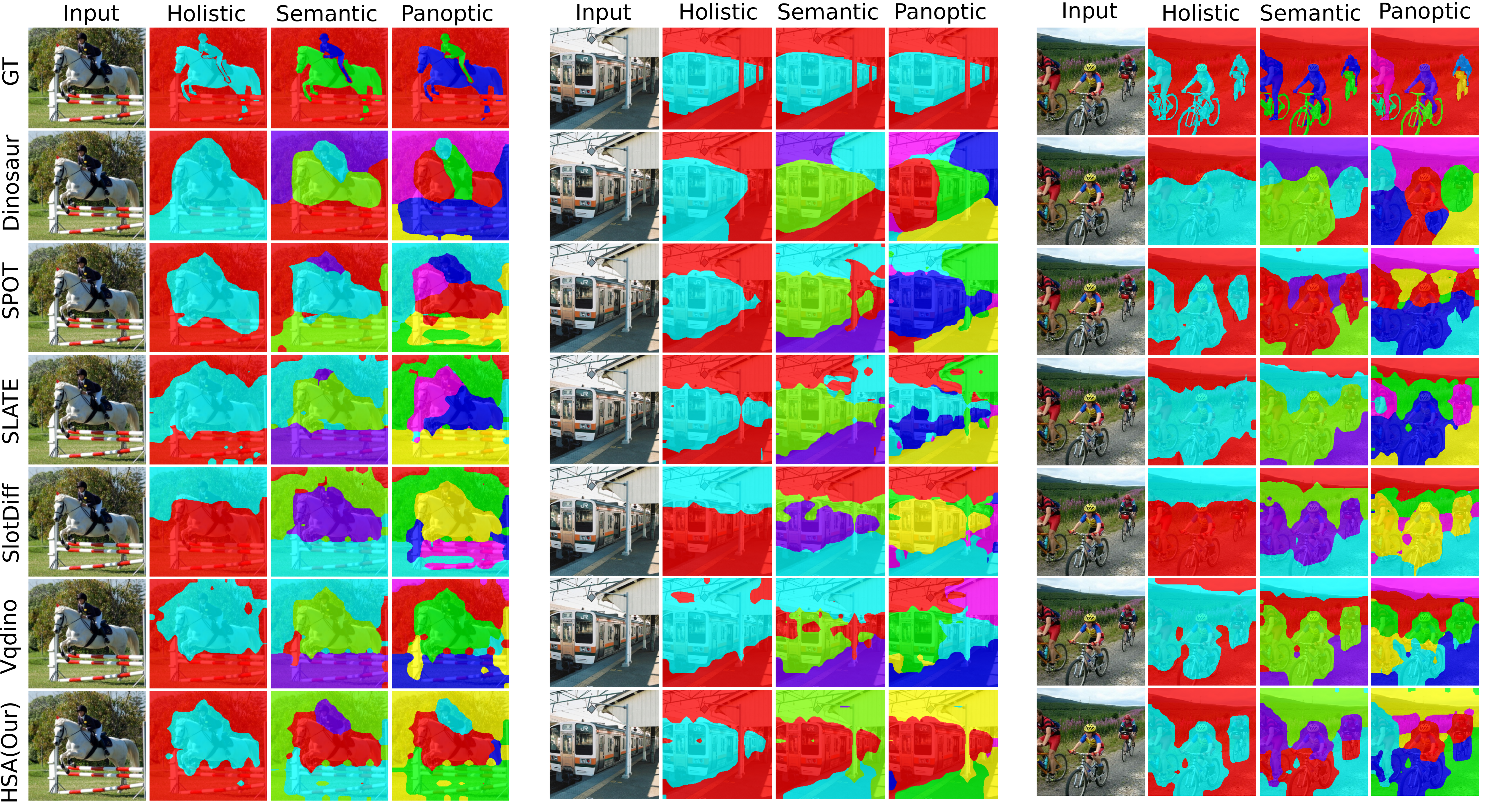}
\caption{Qualitative comparison on VOC dataset. \textbf{Rows 2--6}: flat baselines (DINOSAUR, SPOT, SLATE, SlotDiffusion, VQDINO) produce semantically arbitrary decompositions. \textbf{Row 7 (HSA, ours)}: cleanly separate objects, and are semantically grounded. Best viewed in color.}
\vspace{-0.3cm}
\label{fig:qualitative_voc_supp}
\end{figure*}

\noindent \textbf{Instance-Level Inference Across Hierarchy Levels.}
We evaluate instance-level discovery in Table~\ref{tab:instance_levels} by running $K=7$ slot inference using the aggregator weights from each of HSA's three hierarchy levels. The holistic aggregator achieves high overall ARI (59.8\%) because its coarse foreground/background training causes it to subdivide the foreground region into large clusters that align well with global ground-truth groupings at the pixel level. However, its low FG-ARI (15.9\%) reveals that these clusters do not correspond to clean instance boundaries; the holistic aggregator was never trained to separate individual objects, meaning its subdivisions are spatially consistent but semantically imprecise. 

The semantic aggregator dramatically improves FG-ARI to 42.1\%, reflecting significantly better category-level grouping. The panoptic aggregator achieves the best overall performance on object metrics, with peak FG-ARI (42.7\%), mBO (36.9\%), and mIoU (35.3\%), as it was explicitly optimized for fine-grained instance decomposition. This ordered improvement across levels (holistic $\rightarrow$ semantic $\rightarrow$ panoptic) in FG-ARI, mBO, and mIoU confirms that each HSA level specializes for its designated granularity and that the learned hierarchy encodes semantically meaningful structure. This motivates our choice of the panoptic level for instance-level inference in the main paper.

\section{Additional Qualitative results}

This section shows and discusses additional qualitative results on COCO val2017~\cite{Lin-ECCV-2014}, and Pascal VOC~\cite{everingham2010pascal} datasets. \\

\noindent \textbf{Results on COCO val 2017.}
Figure~\ref{fig:qualitative_coco_supp} shows additional qualitative results on COCO val2017~\cite{Lin-ECCV-2014}. Rows 1--6 show results from GT, DINOSAUR~\cite{Seitzer-ICLR-2022}, SPOT~\cite{kakogeorgiou2024spot}, SLATE~\cite{singh2022illiterate}, SlotDiffusion~\cite{wu2023slotdiffusion}, and VQDiNo~\cite{zhao2025vector} respectively; the last row shows HSA (ours). HSA produces cleaner and more coherent decompositions at all three levels compared to flat baselines. The effect is most evident at the holistic level, where HSA cleanly separates foreground from background, while baselines produce fragmented or incoherent splits. At semantic and panoptic levels, flat baselines suffer from over-clustering, a known failure mode in OCL~\cite{Locatello-NeurIPS-2020} where increasing slot count leads to arbitrary fragmentation rather than meaningful decomposition. HSA maintains coherent decompositions across all three levels simultaneously. Notably, in the middle example (mug and knife), the GT annotates the entire scene as foreground, while HSA separately discovers the mug and knife as distinct objects. This demonstrates that HSA retains the object discovery capability of unsupervised OCL models while adding semantic grounding.

\noindent \textbf{Results on Pascal VOC.}
Figure~\ref{fig:qualitative_voc_supp} shows additional qualitative results on Pascal VOC~\cite{everingham2010pascal}. As in COCO, flat baselines produce semantically arbitrary decompositions, with over-clustering becoming more severe at finer granularities. HSA maintains coherent decompositions across all three levels. In the first example (horse jumping), HSA correctly separates the horse, rider, and barrier at the semantic level, including the barrier as a distinct object not separately annotated in GT, again demonstrating HSA's object discovery capability beyond ground-truth labels. In the 
second example (train), HSA cleanly decomposes the large uniform train surface into meaningful regions, while baselines such as SLATE and VQDiNo produce arbitrary fragmentation. The third example (cyclists) best illustrates HSA's semantic grouping: the semantic level groups all cyclists under one slot and all bicycles under another, while the panoptic level separates individual cyclist-bicycle pairs. This is precisely the kind of category-then-instance hierarchy that flat baselines cannot achieve without cross-level 
joint training.